\documentclass{bmvc2k}
\usepackage{hyperref}
\usepackage{amsmath,amsfonts,amssymb}
\usepackage{graphicx}

\title{Multi-task Mid-level Feature Alignment Network for Unsupervised Cross-Dataset Person Re-Identification}

\addauthor{Shan Lin}{shan.lin@warwick.ac.uk}{1}
\addauthor{Haoliang Li}{lihaoliang@ntu.edu.sg}{2}
\addauthor{Chang-Tsun Li}{chli@csu.edu.au}{3}
\addauthor{Alex Chichung Kot}{eackot@ntu.edu.sg}{2}

\addinstitution{
 Department of Computer Science\\
 University of Warwick\\
 United Kingdom
}
\addinstitution{
 Rapid-Rich Object Search Lab\\
 Nanyang Technological University\\
 Singapore
}

\addinstitution{
 School of Computing \& Mathematics\\
 Charles Sturt University\\
 Australia
}

\runninghead{Lin \etal}{Mid-level Feature Alignment for Unsupervised Person Re-ID}

\def\etal{\emph{et al}\bmvaOneDot}

\begin{document}
\maketitle

\begin{abstract}
Most existing person re-identification (Re-ID) approaches follow a supervised learning framework, in which a large number of labelled matching pairs are required for training. Such a setting severely limits their scalability in real-world applications where no labelled samples are available during the training phase. To overcome this limitation, we develop a novel unsupervised \textbf{M}ulti-task \textbf{M}id-level \textbf{F}eature \textbf{A}lignment (MMFA) network for the unsupervised cross-dataset person re-identification task. Under the assumption that the source and target datasets share the same set of mid-level semantic attributes, our proposed model can be jointly optimised under the person's identity classification and the attribute learning task with a cross-dataset mid-level feature alignment regularisation term. In this way, the learned feature representation can be better generalised from one dataset to another which further improve the person re-identification accuracy. Experimental results on four benchmark datasets demonstrate that our proposed method outperforms the state-of-the-art baselines. 
\end{abstract}

\section{Introduction}
Person Re-identification (Re-ID) is the problem of identifying the re-appearing person in a non-overlapping multi-camera surveillance system. Two primary tasks in person Re-ID are learning the subjects' features and developing new similarity measurements, which should be invariant to the viewpoint, pose, illumination and occlusion. Due to its potential applications in security and surveillance, person Re-ID has received substantial attention from both academia and industry. As a result, the person Re-ID performance on existing datasets has been significantly improved in the recent years. For example, the Rank-1 accuracy of a single query search on the Market1501 dataset \cite{Zheng2015ScalableBenchmark} has been pushed from 44.4\% \cite{Zheng2015ScalableBenchmark} to 91.2\% \cite{Li2018HarmoniousRe-Identification}. The Rank-1 accuracy of the DukeMTMC-reID dataset \cite{Zheng2017UnlabeledVitro} released in 2017 has been quickly improved from 30.8\% \cite{Liao2015PersonLearning} to 81.8\%\cite{Si2018DualRe-Identification}. However, most of these approaches follow supervised learning frameworks which required a large number of manually labelled images
. In real-world person Re-ID deployment, typical video surveillance systems usually consist of over one hundred cameras. Manual labelling all those cameras is a prohibitively expensive job. The limited scalability severely hinders the applicability of existing supervised Re-ID approaches in the real-world scenarios.

One solution to make a person Re-ID model scalable is designing an unsupervised algorithm for the unlabelled data. In recent years, some unsupervised methods have been proposed to extract view-invariant features and measure the similarity of images without label information \cite{Wang2016TowardsRe-Identification,Kodirov2015DictionaryRe-identification,Wang2014UnsupervisedRe-identification,Yu2017Cross-ViewRe-Identification}. These approaches only analyse the unlabelled datasets and generally yield poor person Re-ID performance due to the lack of strong supervised tuning and optimisation. Another approach to solve the scalability issue of Re-ID is unsupervised transfer learning via domain adaptation strategy. The unsupervised domain adaptation methods leverage labelled data in one or more related source datasets (also known as source domains) to learn models for unlabelled data in a target domain. However, most domain adaptation frameworks\cite{Long2015LearningNetworks,Long2017DeepNetworks} assume that the source domain and target domain contain the same set of class labels. Such assumption does not hold for person Re-ID because different Re-ID datasets usually contain completely different sets of persons (classes). Therefore, most unsupervised cross-dataset Re-ID methods proposed in recent years \cite{Peng2016UnsupervisedRe-identification,Wang2018TransferableRe-Identification,Deng2018Image-ImageRe-identification} did not use conventional domain adaptation mechanisms. For example, \cite{Deng2018Image-ImageRe-identification} uses image-to-image translation to transfer the style of images in the target domain to the source domain images for generating a new training dataset. These newly generated samples which inherit the identity labels from the source domain and the image style of the target domain can be used for supervised person Re-ID learning. \cite{Wang2018TransferableRe-Identification}
trains two individual models: identity classification and attribute recognition and performs the domain adaptation between two models.

In our work, we rethink the assumption made for the unsupervised cross-dataset Re-ID. Although the identity labels of the source and target datasets are non-overlapping, many of the mid-level semantic features of different people such as genders, age-groups or colour/texture of the outfits are commonly shared between different people across different datasets. Hence, these mid-level visual attributes of the people can be considered as the common labels between different datasets. If we assume these mid-level semantic features are shared between the different domains, we can then treat the unsupervised cross-dataset person Re-ID as a domain adaptation transfer learning based on the mid-level semantic features from the source domain to the target domain. Therefore, we propose a \textbf{M}ulti-task \textbf{M}id-level \textbf{F}eature \textbf{A}lignment network (MMFA) which can simultaneously learn the feature representation from the source dataset and perform domain adaptation to the target dataset via aligning the distributions of the mid-level features. The contributions of our MMFA model are summarized below:
\begin{itemize}
\item We propose a novel unsupervised cross-dataset domain adaptation framework for person Re-ID by minimising the distribution variation of the source's and the target's mid-level features based on the Maximum Mean Discrepancy (MMD) distance \cite{Gretton2009ATest}. Due to the low dimensionality of attribute annotations, we also include mid-level feature maps in our deep neuron network as additional latent attributes to capture a more completed representation of mid-level features of each domain. In our experiments, the proposed MMFA method surpasses other state-of-the-art unsupervised models on four popular unsupervised benchmarks datasets.

\item The existing unsupervised domain adaptation Re-ID approaches based on deep learning \cite{Wang2018TransferableRe-Identification,Deng2018Image-ImageRe-identification} require two-stage learning processes: supervised feature learning and unsupervised domain adaptation. Different from those methods, Our MMFA model introduce a new jointly training structure which simultaneously learns the feature representation from the source domain and adapts the feature to the target domain in a single training process. Because our model does not require two-step training procedure, the training time for our method is much less than many other unsupervised deep learning person Re-ID approaches.
\end{itemize}

\section{Related Work}
Most existing Re-ID models are supervised approaches focusing on features engineering \cite{Gray2008ViewpointFeatures,Liao2015PersonLearning,Zhao2014LearningRe-identification,Yan2016CNNComplementary}, distance metrics development \cite{Kostinger2012LargeConstraints,Paisitkriangkrai2015LearningEnsembles,Liao2015PersonLearning,Yan2017ExploitingVehicles} or creating new deep learning architectures \cite{Ahmed2015AnRe-Identification,Lin2017End-to-EndRe-Identification,Li2018HarmoniousRe-Identification}. However, in real-world person Re-ID deployment, supervised methods suffer from poor scalability due to the lack of subject's identity for each camera pair. Therefore, some unsupervised person Re-ID methods have been developed based on hand-crafted features learned from a single unlabelled dataset \cite{Zhao2013UnsupervisedRe-identification,Wang2014UnsupervisedRe-identification,Kodirov2015DictionaryRe-identification}. However, due to the absence of the pairwise identity labels, these unsupervised methods cannot learn robust cross-view discriminative features and usually yield much weaker performance compared to the supervised learning approaches. 

Because of the poor person Re-ID performance of the single dataset unsupervised learning, many of recent works are focusing on developing cross-dataset domain adaptation methods for a scalable person Re-ID system \cite{Layne2013DomainRe-identification,Ma2015Cross-DomainSVMs,Peng2016UnsupervisedRe-identification,Wang2018TransferableRe-Identification}. These approaches leverage the pre-trained supervised Re-ID models and adapt these models to the target dataset. Early proposed cross-dataset person Re-ID domain adaptation approaches rely on weak label information in target dataset \cite{Layne2013DomainRe-identification,Ma2015Cross-DomainSVMs}. Therefore, these methods can only be considered as semi-supervised or weakly-supervised learning. The recent cross-dataset works such as UMDL \cite{Peng2016UnsupervisedRe-identification}, SPGAN \cite{Deng2018Image-ImageRe-identification} and TJ-AIDL \cite{Wang2018TransferableRe-Identification} do not require any labelled information from the target dataset and can be considered as fully unsupervised cross-dataset domain adaptation learning. The UMDL method tries to transfer the view-invariant feature representation via multi-task dictionary learning on both source and target datasets. The SPGAN approach uses the generative adversarial network (GAN) to generate new training dataset by transferring the image style from the target dataset to the source dataset while preserving the source identity information. Hence, the supervised training on the new translated dataset can be automatically adapted to the target domain. The TJ-AIDL approach individually trains two models: an identity classification model and an attribute recognition model. The domain adaptation in TJ-AIDL is achieved by minimising the distance between inferred attributes from the identity classification model and the predicted attributes from the attribute recognition model. Compared to the previous single dataset unsupervised approaches, the recent cross-dataset unsupervised domain adaptation methods yield much better performance. Our work improved upon these cross-dataset unsupervised methods by introducing a more intuitive domain adaptation mechanism and proposing a novel jointly training framework for simultaneous feature learning and domain adaptation.

\section{The Proposed Methodology}
One basic assumption behind domain adaptation is that there exists a feature space which underlying both the source and the target domain. Although high-level information like person's identity is not shared between different Re-ID datasets, the mid-level features such as visual attributes can be overlapped between persons. For example, the people in dataset A and dataset B can be different, but some of mid-level semantic information like genders, age-groups, colour of clothes or accessories may be the same. Hence, in our proposed method MMFA, we assume that the source and target datasets contain the same set of mid-level attribute labels. As a result, the unsupervised cross-dataset person Re-ID can be transformed into an unsupervised domain adaptation problem by regularising the distribution variance of the attribute feature space between the source domain and the target domain. 

\begin{figure}[t]
\centering
\begin{minipage}[t]{0.28\textwidth}
\includegraphics[trim=5.5cm 1.3cm 5.5cm 1.3cm,clip,width=0.48\textwidth]{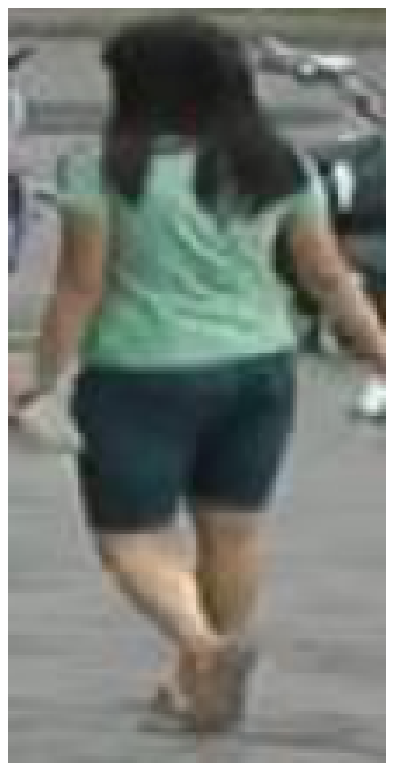}
\includegraphics[trim=5.5cm 1.3cm 5.5cm 1.3cm,clip,width=0.48\textwidth]{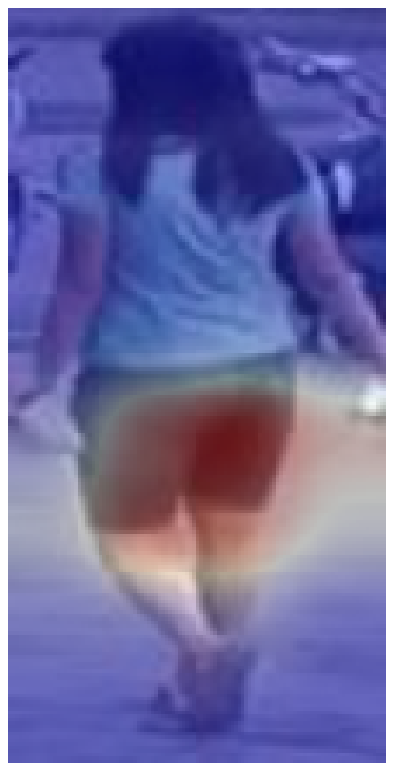}
\small \centering (a) Person ID 0585
\end{minipage}
\hspace{1em}
\begin{minipage}[t]{0.28\textwidth}
\includegraphics[trim=5.5cm 1.3cm 5.5cm 1.3cm,clip,width=0.48\textwidth]{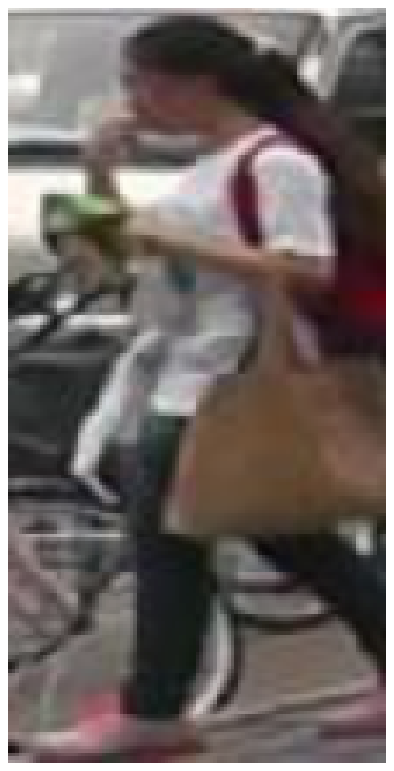}
\includegraphics[trim=5.5cm 1.3cm 5.5cm 1.3cm,clip,width=0.48\textwidth]{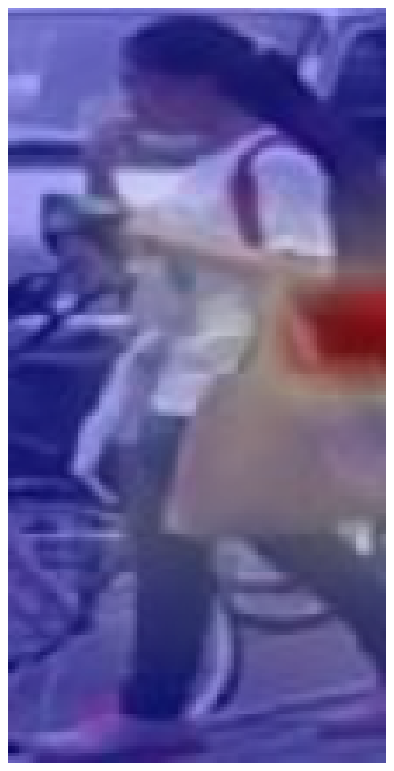}
\small \centering (b) Person ID 0646
\end{minipage}
\hspace{1em}
\begin{minipage}[t]{0.28\textwidth}
\includegraphics[trim=5.5cm 1.3cm 5.5cm 1.3cm,clip,width=0.48\textwidth]{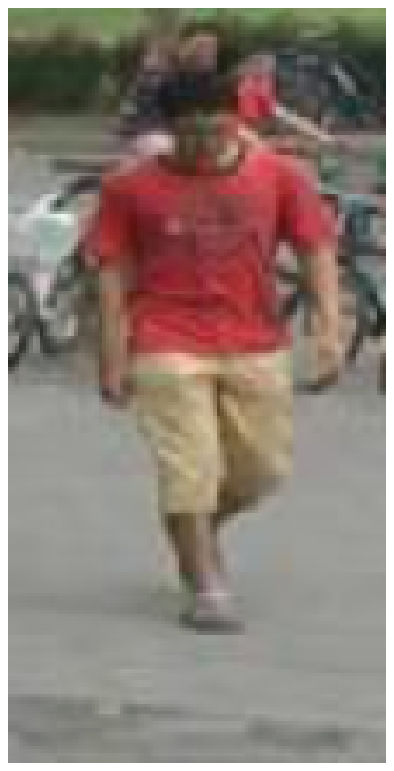}
\includegraphics[trim=5.5cm 1.3cm 5.5cm 1.3cm,clip,width=0.48\textwidth]{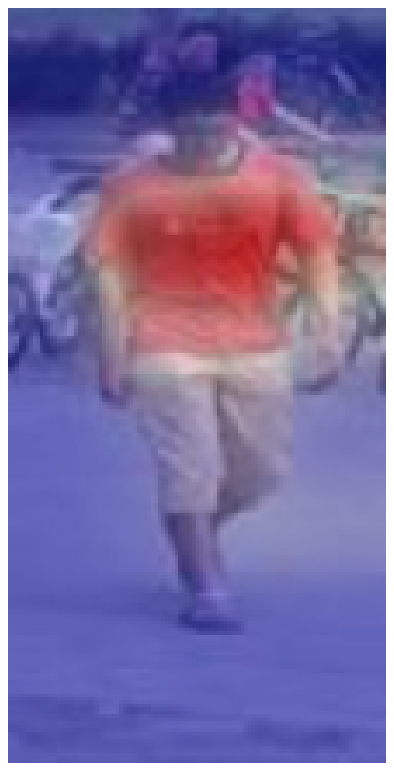}
\small \centering (c) Person ID 1091
\end{minipage}
\vspace{0.5em}
\caption{In each of these three pairs of images, the one on the left-hand side is randomly selected from the Market1501 dataset while the other one shows the attention regions from highest activated feature maps ($1749_{th}$, $511_{th}$ and $1091_{th}$) of the last convoltional layer. These feature maps highlight distinctive semantic features such as green shorts, red backpack, red T-shirt. Best view in colour.}
\label{fig:feature-map}
\end{figure}

Currently, there are a few attribute annotations available for some Re-ID datasets. However, the number of these attribute labels are limited. There are 27 attribute labels for the Market1501 dataset and 23 for the DukeMTMC-reID dataset \cite{Lin2017ImprovingLearning}. The features from 27 or 23 user-defined attributes alone cannot give a good representation of the overall mid-level semantic features for both source and target datasets. There may exist many shared mid-level visual clues between domains which cannot fully captured by those 27/23 user-defined annotations. To obtain more attributes for representing the shared mid-level features, we start to consider the feature-maps generated from the different convolutional layers. In our experiment, we observed that most feature maps from the last convolutional layer of an attribute-identity multi-task classification model are able to capture many distinctive semantic features of a person,  see Figure \ref{fig:feature-map} for example. Hence, we treat those feature maps as the attribute-like mid-level deep features in our proposed MMFA model.  

\subsection{Architecture}
\begin{figure}[t]
\centering
\includegraphics[trim=0cm 0.7cm 0cm 0cm,clip, width=0.95\textwidth]{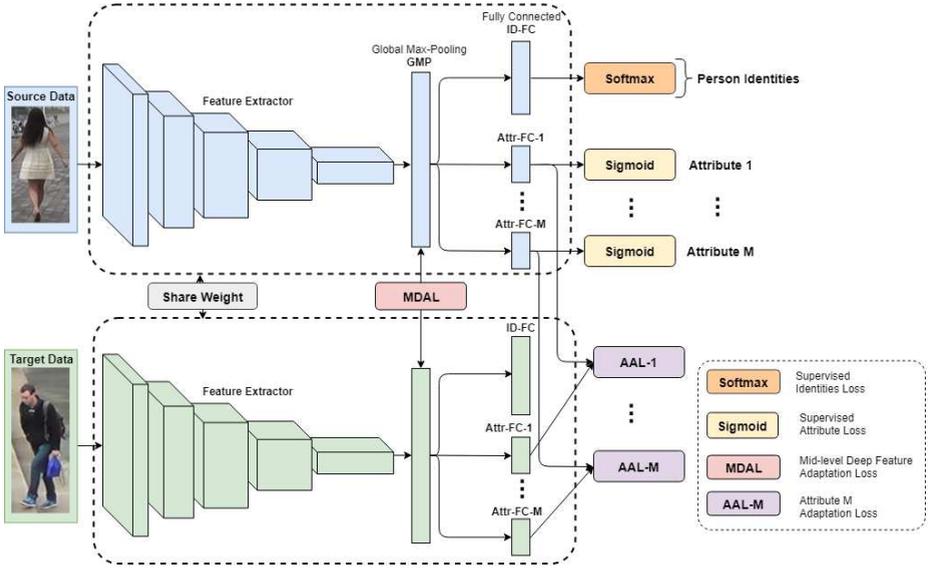}
\caption{In the proposed MMFA model, the source and target images will undergo two networks with shared weights. The global max-pooling will extract the most sensitive feature maps from the last convolutional layer and feed them into each independent softmax classifier for classifying the identity or attributes of the person. In order to generalise the feature representation to the target dataset, we also regularised our network by aligning the distribution of the pre-defined attributes and mid-level deep features from the source to the target domain.}
\label{fig:architecture}
\end{figure}
Our model is optimised using stochastic gradient descent (SGD) method on mini-batches. Each mini-batch consists $n_S$ number of labelled images $[\mathbf{I}_{S,1},\mathbf{I}_{S,2},...,\mathbf{I}_{S,{n_S}}]$ from a source dataset $S$ and $n_T$ number of unlabelled images $ [\mathbf{I}_{T,1},\mathbf{I}_{T,2},...,\mathbf{I}_{T,{n_T}}]$ from a target dataset $T$. Each labelled image $\mathbf{I}_{S,i}$ is associated with an identity label $y_{S,i}$ and a set of M attributes $\mathbf{A}_{S,i} = [a_{S,i}^1,a_{S,i}^2,...,a_{S,i}^M]$. Our model consists of one pre-trained ResNet50-based backbone network \cite{He2016DeepRecognition} as the feature extractor with 1 fully connected layer for identity classification and $M$ individual fully connected layers for single attribute recognition. The overview of our architecture is shown in Figure \ref{fig:architecture}. We change the last average pooling layer from ResNet50 to a global max-pooling (GMP) layer to emphasise the semantic regions from the feature maps of the last convolutional layer. 

$\mathbf{H}_S = [\mathbf{h}_{S,1},\mathbf{h}_{S,2},...,\mathbf{h}_{S,{n_S}}]$ and $\mathbf{H}_T = [\mathbf{h}_{T,1},\mathbf{h}_{T,2},...,\mathbf{h}_{T,{n_T}}]$ are the mid-level deep features of the source domain and the target domain obtained after the GMP layer, respectively. The identity features $\mathbf{H}_S^{id} = [\mathbf{h}_{S,1}^{id},\mathbf{h}_{S,2}^{id},...,\mathbf{h}_{S_{n,S}}^{id}]$ and $\mathbf{H}_T^{id}= [\mathbf{h}_{T,1}^{id},\mathbf{h}_{T,2}^{id},...,\mathbf{h}_{T,{n_T}}^{id}]$ are the outputs from the fully connected layer with $\mathbf{H}_S$ and $\mathbf{H}_T$ as input for identity classification (shown as \textit{ID-FC}  in Figure \ref{fig:architecture}). For a specific $m$-th attribute where $m\in M$, the $m$-th attribute features $\mathbf{H}^{attr_m}_S = [\mathbf{h}_{S,1}^{attr_m},\mathbf{h}_{S,2}^{attr_m},...,\mathbf{h}_{S,{n_{S}}}^{attr_m}]$, $\mathbf{H}^{attr_m}_T = [\mathbf{h}_{T,1}^{attr_m},\mathbf{h}_{T,2}^{attr_m},...,\mathbf{h}_{T,{n_{T}}}^{attr_m}]$ can be obtained from its corresponding fully connected layer with $\mathbf{H}_S$ and $\mathbf{H}_T$ as input (shown as \textit{Attr-FC-m} in Figure \ref{fig:architecture}). Our model can be jointly trained in a multi-task manner: two supervised classification losses for identity classification and attribute recognition, one adaptation losses based on the attribute features and another adaptation loss based on the mid-level deep features.

\subsection{Multi-task Supervised Classification for Feature Learning}
The view-invariant feature representations are learned from a multi-task identity and attribute classification training. The additional attribute annotations provide further regularisation and additional supervision to the feature learning process

\noindent
\underline{\textbf{Identity Loss: }}
We denote that $p_{id}(\mathbf{h}^{id}_{S,i},y_{S,i})$ is the predicted probability on the identity feature $\mathbf{h}^{id}_{S,i}$ with the ground-truth label $y_{S,i}$. The identity loss is computed by softmax cross entropy function:
\begin{align}
L_{id}=-\frac{1}{n_S}\sum^{n_S}_{i=1}log(p_{id}(\mathbf{h}^{id}_{S,i},y_{S,i}))
\end{align}

\noindent
\underline{\textbf{Attribute Loss: }} 
We denote that $p_{attr}(\mathbf{h}^{attr_m}_{S,i},m)$ is the predicted probability for the $m$-th attribute feature $\mathbf{h}^{attr_m}_{S,i}$ with ground-truth label $a_{S,i}^m$. 
The overall attributes loss can be expressed as the average of sigmoid cross entropy loss of each attribute:
\begin{align}
L_{attr}=-\frac{1}{M}\frac{1}{n_S}\sum^{M}_{m=1}\sum^{n_S}_{i=1}(a_{S,i}^m\cdot log(p_{attr}(\mathbf{h}^{attr_m}_{S,i},m))+(1-a_{S,i}^m)\cdot log(1-p_{attr}(\mathbf{h}^{attr_m}_{S,i},m)))
\end{align}

\subsection{MMD-based Regularisation for Mid-level Feature Alignment}
As we make a shared mid-level latent space assumption in our MMFA model, the domain adaptation can be achieved by reducing the distribution distance of attribute features between the source domain and the target domain. Based on the attribute features $\{\mathbf{H}^{attr_1}_S,..,\mathbf{H}^{attr_M}_S\}$ and $\{\mathbf{H}^{attr_1}_T,..,\mathbf{H}^{attr_M}_T\}$ obtained from the supervised classification learning, we use the Maximum Mean Discrepancy (MMD) measure \cite{Gretton2009ATest} to calculate the feature distribution distance of each attribute. The overall attribute distribution distance is the the mean MMD distance of all attributes:
\begin{align}
L_{AAL} = \frac{1}{M}\sum^{M}_{m=1}MMD(\mathbf{H}^{attr_m}_S,\mathbf{H}^{attr_m}_T)^2 = \frac{1}{M}\sum^{M}_{m=1}\left \| \frac{1}{n_{S}}\sum^{n_{S}}_{i=1}\phi(\mathbf{h}^{attr_m}_{S,i}) - \frac{1}{n_{T}}\sum^{n_{T}}_{j=1}\phi(\mathbf{h}^{attr_m}_{T,j}) \right \|_{\mathcal{H}}^2
\end{align}

\noindent
$\phi(\cdot)$ is a map operation which project the attribute distribution into a reproducing kernel Hilbert space (RKHS) $\mathcal{H}$ \cite{Gretton2008AProblem}. $n_S$ and $n_T$ are the batch sizes of the source domain images and target domain images. The arbitrary distribution of the attribute features can be represented by using the kernel embedding technique \cite{Smola2007ADistributions}. It has been proven that if the kernel $k(\cdot,\cdot)$ is characteristic, then the mapping to the RKHS $\mathcal{H}$ is injective \cite{Sriperumbudur2009KernelDistributions}. The injectivity indicates that the arbitrary probability distribution is uniquely represented by an element in RKHS. Therefore, we have a kernel function $k(\mathbf{h}^{attr_m}_{S,i},\mathbf{h}^{attr_m}_{T,j})=\phi(\mathbf{h}^{attr_m}_{S,i})\phi(\mathbf{h}^{attr_m}_{T,j})^\intercal$ induced by $\phi(\cdot)$. Now, the average MMD distance between the source domain's and the target domain's attribute distributions can be re-expressed as:
\begin{align}
L_{AAL} = \frac{1}{M}\sum^{M}_{m=1}\Big[&\frac{1}{({n_S})^2}\sum^{n_S}_{i=1}\sum^{n_S}_{i'=1}k(\mathbf{h}^{attr_m}_{S,i},\mathbf{h}^{attr_m}_{S,{i'}}) +\frac{1}{({n_T})^2}\sum^{n_T}_{j=1}\sum^{n_T}_{j'=1}k(\mathbf{h}^{attr_m}_{T,j},\mathbf{h}_{T,{j'}}^{attr_m})\nonumber \\
&-\frac{2}{n_S\cdot n_T}\sum^{n_S}_{i=1}\sum^{n_T}_{j=1}k(\mathbf{h}_{S,i}^{attr_m},\mathbf{h}_{T,j}^{attr_m})\Big]
\end{align}

\noindent
In our MMFA model, we decided to use the well-know RBF characteristic kernel with bandwidth $\alpha$: 
\begin{align}
k(\mathbf{h}^{attr_m}_{S,i},\mathbf{h}^{attr_m}_{T,j})=exp(-\frac{1}{2\alpha}\left \| \mathbf{h}^{attr_m}_{S,i} - \mathbf{h}^{attr_m}_{T,j} \right \|^2) 
\end{align}

\noindent
Due to the limited size of available attribute annotations, these attributes alone cannot give a good representation of all domain shared mid-level features. By assuming the last feature maps after the feature extractor are attribute-like mid-level features, we introduce the additional mid-level deep feature alignment to our model. The mid-level deep features adaptation loss $L_{MDAL}$ is the MMD distance between the source and target mid-level deep feature $\mathbf{H}_S,\mathbf{H}_T$, similar to our attributes features adaptation loss: 
\begin{align}
L_{MDAL}=MMD(\mathbf{H}_S,\mathbf{H}_T)^2= \left \| \frac{1}{n_S}\sum^{n_S}_{i=1}\phi(\mathbf{h}_{S,i}) - \frac{1}{n_T}\sum^{n_T}_{j=1}\phi(\mathbf{h}_{T,j}) \right \|_{\mathcal{H}}^2
\end{align}

\noindent
Finally, we formulate the overall loss function by incorporating the weighted summation of above components $L_{id}$, $L_{attr}$, $L_{AAL}$ and $L_{MDAL}$:
\begin{align} \label{loss eq}
L_{all} = L_{id} + \lambda_1 L_{attr} + \lambda_2 L_{AAL} + \lambda_3 L_{MDAL}
\end{align}

\section{Experiments}
\subsection{Datasets and Settings}
\underline{\textbf{Person Re-ID Datasets:}} Four widely used person Re-ID benchmarks are chosen for experimental evaluations: Market1501, DukeMTMC-reID, VIPeR and PRID. The Market1501 dataset \cite{Zheng2015ScalableBenchmark} contains 32,668 images of 1,501 pedestrian. 751 identities are selected for training and 750 remaining identities are for testing. Each identity was captured by at most 6 non-overlapping cameras. The DukeMTMC-reID dataset \cite{Zheng2017UnlabeledVitro} is the redesign version of pedestrian tracking dataset DukeMTMC \cite{Ristani2016PerformanceTracking} for person Re-ID task. It contains 34,183 image of 1,404 pedestrians. 702 identities are used for training and the remaining 702 are for testing. Each identity was captured by 8 non-overlapping cameras. The VIPeR dataset \cite{Gray2007EvaluatingTracking} is one of the oldest person Re-ID dataset. It contains 632 identities, but only two images for each identity. Due to its low resolution and large variation in illumination and viewpoints, the VIPeR dataset is still a very challenging dataset. The PRID dataset \cite{Hirzer2011PersonClassification} consists of 934 identities from two camera views. There are 385 identities in View A and 749 identities in View B, but only 200 identities appear in both views.

\noindent
\underline{\textbf{Evaluation Protocol:}} We follow the proposed single-query evaluation protocols for Market1501 and DukeMTMC-reID \cite{Zheng2015ScalableBenchmark,Zheng2017UnlabeledVitro}. For the VIPeR dataset, we randomly half-split the dataset into training and testing sets. The overall performance on VIPeR is the average results from 10 randomly 50/50 split testing. For the PRID dataset evaluation, we follow the same single-shot experiments as \cite{Zhang2016LearningRe-identification}. Similar to the VIPeR dataset setting, the final performance is the average of the experimental results based on 10 random split testing. Since the VIPeR and PRID datasets are too small for training the deep learning network, our MMFA model only trains on the Market1501 or the DukeMTMC-reID datasets. We adopted the commonly used Cumulative Matching Characteristic (CMC) and mean Average Precision (mAP) as performance metrics.

\noindent
\underline{\textbf{Implementation Details:}} The input images are randomly cropped and resized to (256,128,3). All the fully-connected layers after global max-pooling layer are equipped with batch normalization, the dropout rate of 0.5 and the leaky RELU activation function. $\lambda1$, $\lambda2$ and $\lambda3$ in the final loss function (Equation \ref{loss eq}) are empirically fixed to $0.1,1,1$. For all the adaptation losses, we adopted the mixture kernel strategy \cite{Li2015GenerativeNetworks,Li2018DomainLearning} by averaging the RBF kernels with the bandwidth $\alpha=1, 5, 10$. We use the stochastic gradient descent (SGD) optimizer with batch size 32 for both source domain images and the target domain images. We set the learning rate to $0.01$ and the nesterov momentum to $0.9$ with the weight decay of $5\times10^{-4}$. The learning rate will decrease by 10 after the $20$-th epoch. The person Re-ID evaluation of the target domain is measured by the $L_2$ distance of the 2048-D mid-level deep features $H_T$ after the global max-pooling layer.

\subsection{Comparisons with State-of-the-art Methods}
The performance of our proposed MMFA model is extensively compared with 16 state-of-the-art unsupervised person Re-ID methods as shown in Table \ref{soa-comparison}. These methods include: view-invariant feature learning methods SDALF \cite{Farenzena2010PersonFeatures} and CPS \cite{Cheng2011CustomRe-identification}, graph learning method GL \cite{Kodirov2016PersonLearning}, sparse ranking method ISR \cite{Lisanti2015PersonRanking}, salience learning methods GTS \cite{Wang2014UnsupervisedRe-identification} and SDC \cite{Zhao2017PersonLearning}, neighbourhood clustering methods AML \cite{Ye2007AdaptiveClustering}, UsNCA \cite{Qin2015UnsupervisedClustering}, CAMEL \cite{Yu2017Cross-ViewRe-Identification} and PUL \cite{Fan2017UnsupervisedFine-tuning}, ranking SVM method AdaRSVM \cite{Ma2015Cross-DomainSVMs}, attribute co-training method SSDAL \cite{Su2016DeepRe-identification}, dictionary learning method DLLR \cite{Kodirov2015DictionaryRe-identification} and UDML \cite{Peng2016UnsupervisedRe-identification}, id-to-attribute transfer method TJ-AIDL \cite{Wang2018TransferableRe-Identification} and image style transfer method SPGAN \cite{Deng2018Image-ImageRe-identification}. These methods can be categorised into three groups:
\begin{enumerate} 
\setlength\itemsep{0em}
\item hand-craft features approaches: SDALF,CPS,DLLR,GL,ISR,GTS,SDC
\item clustering approaches: AML, UsNCA, CAMEL, PUL
\item domain adaptation approaches: AdaRSVM, UDML, SSDAL, TJ-AIDL, SPGAN
\end{enumerate}

\begin{table}[t]
\centering
\resizebox{0.70\textwidth}{!}{%
\begin{tabular}{l|c|c|c|c|c|c}
\hline
Dataset & VIPeR & PRID & \multicolumn{2}{c|}{Market1501} & \multicolumn{2}{c}{DukeMCMT-reID} \\ \hline
Metric (\%) & Rank-1 & Rank-1 & Rank-1 & mAP & Rank-1 & mAP \\ \hline
SDALF \cite{Farenzena2010PersonFeatures} & 19.9 & 16.3 & - & - & - & - \\
CPS \cite{Cheng2011CustomRe-identification} & 22.0 & - & - & - & - & - \\
DLLR \cite{Kodirov2015DictionaryRe-identification} & 29.6 & 21.1 & - & - & - & - \\
GL \cite{Kodirov2016PersonLearning} & 33.5 & 25.0 & - & - & - & - \\
ISR \cite{Lisanti2015PersonRanking} & 27.0 & 17.0 & 40.3 & 14.3 & - & - \\
GTS \cite{Wang2014UnsupervisedRe-identification} & 25.2 & - & - & - & - & - \\
SDC \cite{Zhao2017PersonLearning} & 25.8 & - & - & - & - & - \\ \hline
AML \cite{Ye2007AdaptiveClustering} & 23.1 & - & 44.7 & 18.4 & - & - \\
UsNCA \cite{Qin2015UnsupervisedClustering} & 24.3 & - & 45.2 & 18.9 & - & - \\
CAMEL \cite{Yu2017Cross-ViewRe-Identification} & 30.9 & - & 54.5 & 26.3 & - & - \\
PUL \cite{Fan2017UnsupervisedFine-tuning} & - & - & 44.7 & 20.1 & 30.4 & 16.4 \\ \hline
AdaRSVM \cite{Ma2015Cross-DomainSVMs} & 10.9 & 4.9 & - & - & - & - \\
UDML \cite{Peng2016UnsupervisedRe-identification} & 31.5 & 24.2 & - & - & - & - \\
SSDAL \cite{Su2016DeepRe-identification}  & 37.9 & 20.1 & 39.4 & 19.6 & - & - \\
TJ-AIDL\textsuperscript{Duke} \cite{Wang2018TransferableRe-Identification} & 35.1 &\underline{34.8} & \textbf{58.2} & 26.5 & - & - \\
SPGAN\textsuperscript{Duke} \cite{Deng2018Image-ImageRe-identification} &- & - & 51.1 & 22.8 & - & - \\
TJ-AIDL\textsuperscript{Market} \cite{Wang2018TransferableRe-Identification} & \underline{38.5} & 26.8 & - & - & \underline{44.3} & \underline{23.0} \\ 
SPGAN\textsuperscript{Market} \cite{Deng2018Image-ImageRe-identification} & - & - & - & - & 41.1 & 22.3 \\ \hline
\textbf{MMFA\textsuperscript{Duke}} & 36.3 & 34.5 & \underline{56.7} & \textbf{27.4} & - & - \\
\textbf{MMFA\textsuperscript{Market}} & \textbf{39.1} & \textbf{35.1} & - & - & \textbf{45.3} & \textbf{24.7} \\ \hline
\end{tabular}}
\caption{Performance comparisons with state-of-the-art unsupervised person Re-ID methods.The best and second best results are highlighted by bold and underline receptively. The superscripts: \textit{Duke} and \textit{Market} indicate the source dataset which the model is trained on.}
\label{soa-comparison}
\end{table}
Our MMFA method outperforms most existing state-of-the-art models on VIPeR, PRID, Market1501 and DukeMTMC-reID datasets. the Rank-1 accuracy increases from 38.5\% to 39.1\% in VIPeR, from 34.8\% to 35.1\% in PRID and from 44.3\% to 45.3\% in DukeMTMC-reID. The mAP performance of our approach surpasses all exiting methods by a good margin from 23.0\% to 24.7\% and 26.5\% to 27.4\% in DukeMTMC-reID and Market1501 receptively. Although, the Rank-1 accuracy of our MMFA model on the Maket1501 dataset did not surpass the TJ-AIDL method, our mAP score and the overall performance (Rank-5 to Rank-10 accuracy) are better than TJ-AIDL. The complete comparisons with TH-AIDL and SPGAN are shown in Table \ref{detail-comparison}. It is worth noting that the performance of our MMFA is achieved in one single end-to-end training session with only 25 epochs. Our performance can be further improved by implementing any pre- and post-processing techniques such as part-based local max pooling (LMP), attention mechanisms or re-ranking. For fair comparisons, the performance results shown the Table \ref{soa-comparison} and Table \ref{detail-comparison} are all based on the basic models without any pre or post-processing.
\begin{table}[h]
\centering
\resizebox{0.9\textwidth}{!}{%
\begin{tabular}{l|ccc|c|ccc|c}
\hline
Source$\,\to\,$Target & \multicolumn{4}{c|}{Market1501 $\,\to\,$ DukeMTMC-reID} & \multicolumn{4}{c}{DukeMTMC-reID $\,\to\,$ Market1501} \\ \hline
Metric (\%) & Rank1 & Rank5 & Rank10 & mAP & Rank1 & Rank5 & Rank10 & mAP \\ \hline
SPGAN & 41.1 & 56.6 & 63.0 & 22.3 & 51.5 & 70.1 & 76.8 & 22.8 \\
TJ-AIDL & 44.3 & 59.6 & 65.0 & 23.0 & \textbf{58.2}& 74.8 & 81.1 & 26.5 \\
\textbf{MMFA} & \textbf{45.3} & \textbf{59.8} & \textbf{66.3} & \textbf{24.7} & \underline{56.7} & \textbf{75.0} & \textbf{81.8} & \textbf{27.4} \\ \hline
\end{tabular}}
\caption{Detail Comparison with SPGAN and TJ-AIDL}
\label{detail-comparison}
\end{table}

\subsection{Component Analysis and Evaluation}
We also analysed each component of our MMFA model based on their contributions to the cross-domain feature learning. The first set of experiments is the unsupervised performance based on the feature representation learned from the source domain attributes or identity, without any domain adaptation. In the top section of Table \ref{components-analysis}, the attribute annotations alone cannot give a good representation of a person due to its low dimensionality, only 6.4\% and 19.2\% Rank1 accuracy achieved. The features from identity labels on the other hand yield much better performance compared to attributes. When attribute and identity information are jointly trained as a multi-objective learning task, the feature representations show a better generalization-ability. This experiment shows that the attribute annotations do provide extra information to the system which serves as additional supervision for learning more generalised cross-dataset features. 

\begin{table}[h]
\centering
\resizebox{\textwidth}{!}{%
\begin{tabular}{c|ccc|c|ccc|c}
\hline
Source $\,\to\,$ Target & \multicolumn{4}{c|}{Market1501 $\,\to\,$ DukeMTMC-reID} & \multicolumn{4}{c}{DukeMTMC-reID $\,\to\,$ Market1501} \\ \hline
Metric (\%) & Rank1 & Rank5 & Rank10 & mAP & Rank1 & Rank5 & Rank10 & mAP \\ \hline
Attribute Only & 6.4 & 14.4 & 18.6 & 2.3 & 19.2 & 34.8 & 45.1 & 6.2 \\
ID Only & 37.6 & 54.9 & 61.6 & 22.6 & 48.2 & 66.1 & 73.3 & 21.6 \\
Attribute+ID Only & 41.7 & 57.5 & 63.6 & 23.3 & 52.2 & 69.1 & 75.7 & 23.5 \\ \hline
Attribute with Attribute Feature Adaptation & 15.8 & 26.0 & 48.2 & 5.7 & 35.5 & 55.3 & 64.0 & 12.7 \\
ID with Mid-level Deep Feature Adaptation & 42.1 & 57.7 & 63.9 & 24.3 & 53.4 & 70.2 & 76.4 & 25.2 \\
Mid-level Deep Feature + Attribute Adaptation & \textbf{45.3} & \textbf{59.8} & \textbf{66.3} & \textbf{24.7} & \textbf{56.7} & \textbf{75.0} & \textbf{81.8} & \textbf{27.4} \\ \hline
\end{tabular}}
\caption{Adaptation performance on each model components}
\label{components-analysis}
\end{table}
The lower section of Table \ref{components-analysis} shows the unsupervised re-id performances after aligning the mid-level feature distribution. After aligning the source and target distributes of attributes features, mid-level features or both, we can see a large performance increase when compared with previously non-adapted features. It shows that the proposed mid-level feature distribution alignment strategy is a feasible approach for the unsupervised person Re-ID task.

\section{Conclusion}
In this paper, we presented a novel unsupervised cross-dataset feature learning and domain adaptation framework MMFA for person Re-ID task. We utilised the multi-supervision identity and attribute classifications to learn a discriminative feature for person Re-ID on the labelled source dataset. With a shared mid-level feature space assumption, we proposed the mid-level feature alignment domain adaptation strategy to reduce the MMD distance based on the source domain's and the target domain's mid-level feature distributions. In contrast to most existing learn-then-adapt unsupervised cross-dataset approaches, our MMFA is a one-step learn-and-adapt method which can simultaneously learn the feature representation and adapt to the target domain in a single end-to-end training procedure. Meanwhile, our proposed method is still able to outperform a wide range of state-of-the-art unsupervised Re-ID methods.\\

\noindent
\textbf{Acknowledgement}: This work is supported by EU Horizon 2020 project, entitled Computer Vision Enable Multimedia Forensics and People Identification (acronym:IDENTITY, Project ID:690907) and carried out at the Rapid-Rich Object Search (ROSE) Lab at the Nanyang Technological University, Singapore.
\bibliography{reference}
\end{document}